\def\year{2021}\relax
\documentclass[letterpaper]{article} 
\usepackage{aaai21}  
\usepackage{times}  
\usepackage{helvet} 
\usepackage{courier}  
\usepackage[hyphens]{url}  
\usepackage{graphicx} 
\usepackage{natbib}  
\usepackage{caption} 
\usepackage{amsmath, amsfonts, bbm, breqn} 
\usepackage{algorithm, algpseudocode} 
\usepackage{booktabs, multirow, tabularx} 
\newcommand{\emp}[1]{\underline{\textit{\textbf{#1}}}}
\newcommand{\tabincell}[2]{\begin{tabular}{@{}#1@{}}#2\end{tabular}}
\usepackage[switch]{lineno}  %

\usepackage{xcolor}
\usepackage{stfloats}
\usepackage{diagbox}

\title{
Adversarial Training with Fast Gradient Projection Method \\against Synonym Substitution based Text Attacks
}

\author {
        Xiaosen Wang\textsuperscript{\rm 1}\thanks{The first three authors contribute equally.}, 
        Yichen Yang\textsuperscript{\rm 1}\footnotemark[1], 
        Yihe Deng\textsuperscript{\rm 2}\footnotemark[1], 
        Kun He\textsuperscript{\rm 1}\thanks{Corresponding author.} \\
}
\affiliations {
    \textsuperscript{\rm 1} School of Computer Science and Technology,  Huazhong University of Science and Technology \\
    \textsuperscript{\rm 2} Computer Science Department, University of California, Los Angeles \\
    \{xiaosen, yangyc\}@hust.edu.cn, yihedeng@g.ucla.edu, brooklet60@hust.edu.cn
}

\begin{document}

\maketitle

\begin{abstract}
Adversarial training is the most empirically successful approach in improving the robustness of deep neural networks for image classification.
For text classification, however, existing synonym substitution based adversarial attacks are effective but not very efficient to be incorporated into practical text adversarial training. Gradient-based attacks, which are very efficient for images, are hard to be implemented for synonym substitution based text attacks due to the lexical, grammatical and semantic constraints and the discrete text input space. 
Thereby, we propose a fast text adversarial attack method called \textit{Fast Gradient Projection Method (FGPM)} based on synonym substitution, which is about 20 times faster than existing text attack methods and could achieve similar attack performance. 
We then incorporate FGPM with adversarial training and propose a text defense method called \textit{Adversarial Training with FGPM enhanced by Logit pairing} (ATFL). 
Experiments show that ATFL could significantly improve the model robustness and block the transferability of adversarial examples.
\end{abstract}

\section{Introduction}

Deep Neural Networks (DNNs) have 
garnered tremendous success over recent  years~\cite{Alex2012Alexnet,kim2014convolutional,devlin2018bert}. However, researchers also find that DNNs are often vulnerable to \textit{adversarial examples} for image data \cite{Szegedy2014Intriguing} as well as text data \cite{papernot2016crafting}. 
For image classification, numerous methods have been proposed with regard to adversarial attack~\cite{Goodfellow:explaining,wang2019gan} and defense~\cite{Goodfellow:explaining,Chuan2018Countering}. Among which, adversarial training that adopts perturbed examples in the training stage so as to promote the model robustness has become very popular and effective~\cite{Anish2018obfuscated}. 

For natural language processing tasks, however, the lexical, grammatical and semantic constraints and the discrete input space make it much harder to craft text adversarial examples. Current attack methods include character-level attack \cite{liang2017deep,li2018textbugger,Ebrahimi:hotflip}, word-level attack \cite{papernot2016crafting,samanta2017crafting,gong2018adversarial,cheng2018seq2sick,kuleshov:GSA,neekhara2018adversarial,ren:PWWS,wang:IGA}, and sentence-level attack \cite{iyyer2018adversarial,ribeiro2018semantically}. For character-level attack, recent works~\cite{pruthi2019combating} have shown that a spell checker can easily fix the perturbations. 
Sentence-level attacks, usually based on paraphrasing, demand longer time in adversary generation. 
For word-level attack, replacing word based on embedding perturbation or adding/removing word is often vulnerable to the problem of hurting semantic consistency and grammatical correctness. Synonym substitution based attacks could better cope with the above issues and produce adversarial examples that are harder to be detected by humans. Unfortunately, synonym substitution based attacks exhibit lower efficiency compared with existing image attack methods.


As text adversarial attack has attracted increasing interests very recently since 2018, its counterpart, text adversarial defense, is much less studied in the literature. Some research \cite{jia2019certified,huang2019achieving} is based on interval bound propagation (IBP), originally proposed for images~\cite{Gowal2019IBP}, to ensure certified text defense. \citet{zhou2019learning} learn to discriminate perturbations (DISP) and restore the embedding of the original word for defense without altering the training process or the model structure. \citet{wang:IGA} propose a Synonym Encoding Method (SEM), which inserts an encoder before the input layer to defend synonym substitution based attacks. 


To our knowledge, adversarial training, one of the most efficacious defense methods for image data \cite{Anish2018obfuscated}, has not been implemented as an effective defense method against synonym substitution based attacks due to the inefficiency of current adversary generation methods.
On one hand, existing synonym substitution based attack methods
are usually much less efficient to be incorporated into adversarial training. On the other hand, although gradient-based image attacks often have much higher efficiency, it is challenging to adapt such methods directly in the text embedding space to generate meaningful adversarial examples without changing the original semantics, due to the discreteness of the text input space. 

To this end, we propose a gradient-based adversarial attack, called \textit{Fast Gradient Projection Method} (FGPM), for efficient synonym substitution based text adversary generation. Specifically, we approximate the classification confidence change caused by synonym substitution by the product of gradient magnitude and projected distance between the original word and the candidate word in the gradient direction. At each iteration, we substitute a word with its synonym that leads to the highest product value. Compared with existing query-based attack methods, FGPM only needs to calculate the back-propagation once 
to obtain the gradient so as to find the best synonym for each word. Extensive experiments
show that FGPM is about 20 times faster than the current fastest text adversarial attack, and it can achieve similar attack performance and transferability compared with state-of-the-art synonym substitution based adversarial attacks.


With such high efficiency of FGPM, we propose \textit{Adversarial Training with FGPM enhanced by Logit pairing (ATFL)} as an efficient and effective text defense method. Experiments show that ATFL promotes the model robustness against white-box as well as black-box attacks, effectively blocks the transferability of adversarial examples and achieves better generalization on benign data than other defense methods.
Besides, we also find some recent proposed variants of adversarial training for images, such as TRADES \citep{Zhang2019Theoretically}, MMA \citep{Ding2020MMA} that exhibit great effectiveness for image data, cannot improve the performance of adversarial training for text data, indicating the intrinsic difference between text defense and image defense.

\section{Related Work}

This section provides a brief overview on word-level text adversarial attacks and defenses. 

\subsection{Adversarial Attack}
Adversarial attacks fall in two settings: (a) \textit{white-box attack} allows full access to the target model, including model outputs, (hyper-)parameters, gradients and architectures, etc. (b) \textit{black-box attack} only allows access to the model outputs.

Methods based on word embedding usually fall in the white-box setting. 
\citet{papernot2016crafting} find a word in dictionary such that the sign of the difference between the found word and the original word is closest to the sign of the gradient.
However, such word does not necessarily preserve the semantic as well as syntactic correctness and consistency. \citet{gong2018adversarial} further employ the Word Mover’s Distance (WMD) in an attempt to preserve semantics. \citet{cheng2018seq2sick} also propose an attack based on the embedding space with additional constraints targeting seq2seq models.

In black-box setting, \citet{kuleshov:GSA} propose a \textit{Greedy Search Attack (GSA)} that perturbs the input by synonym substitution. Specifically, GSA greedily finds a synonym for replacement that minimizes the classification confidence. \citet{ren:PWWS} propose a \textit{Probability Weighted Word Saliency (PWWS)} that greedily substitutes each target word with a synonym determined by the combination of classification confidence change and word saliency. \citet{alzantot2018generating} also use synonym substitution and propose a population-based algorithm called \textit{Genetic Algorithm (GA)}. \citet{wang:IGA} further propose an \textit{Improved Genetic Algorithm (IGA)} that allows to substitute words in the same position more than once and outperforms GA. 

Our work produces efficient gradient based white-box attacks, while guaranteeing the quality of adversarial examples by restricting the perturbation to synonym substitution, which only appears in black-box attacks. 

\subsection{Adversarial Defense}

There are a series of works~\cite{miyato2016adversarial,sato2018interpretable,barham2019interpretable} that perturb the word embeddings 
and utilize the perturbations for adversarial training as a regularization strategy. These works aim to improve the model performance on the original dataset, but do not intend to defend adversarial attacks. Thus, we do not take such works into consideration.

A stream of recent popular defense methods \cite{jia2019certified,huang2019achieving} focuses on verifiable robustness. They use IBP to train models that are provably robust to all possible perturbations within the constraints. Such endeavor, however, is currently time consuming in the training stage as the authors have noted \cite{jia2019certified} and hard to be scaled to relatively complex models or large datasets. \citet{zhou2019learning} train a perturbation discriminator that validates how likely a token in the text is perturbed and an embedding estimator that restores the embedding of the original word to block adversarial attacks. 

\citet{alzantot2018generating} and \citet{ren:PWWS} adopt the adversarial examples generated by their attack methods for adversarial training and achieve some robustness improvement. 
Unfortunately, due to the relatively low efficiency of adversary generation, they are unable to craft plenty of perturbations during the training to ensure significant robustness improvement. To our knowledge, word-level adversarial training has not been practically applied for text classification as an efficient and effective defense method. 

Besides, \citet{wang:IGA} propose \textit{Synonym Encoding Method (SEM)} that uses a synonym encoder to map all the synonyms to the same code in the embedding space and force the classification to be smoother. Trained with the encoder, their models obtain significant improvement on the robustness with a little decay on the model generalization. 

Different from current defenses, our work focuses on fast adversary generation and easy-to-apply defense method for complex neural networks and large datasets. 

\section{Fast Gradient Projection Method}
In this section, we formalize the definition of adversarial examples for text classification and describe in detail the proposed adversarial attack method, \textit{Fast Gradient Projection Method (FGPM)}.

\begin{figure}[tb]
    \centering
    \includegraphics[scale=0.1]{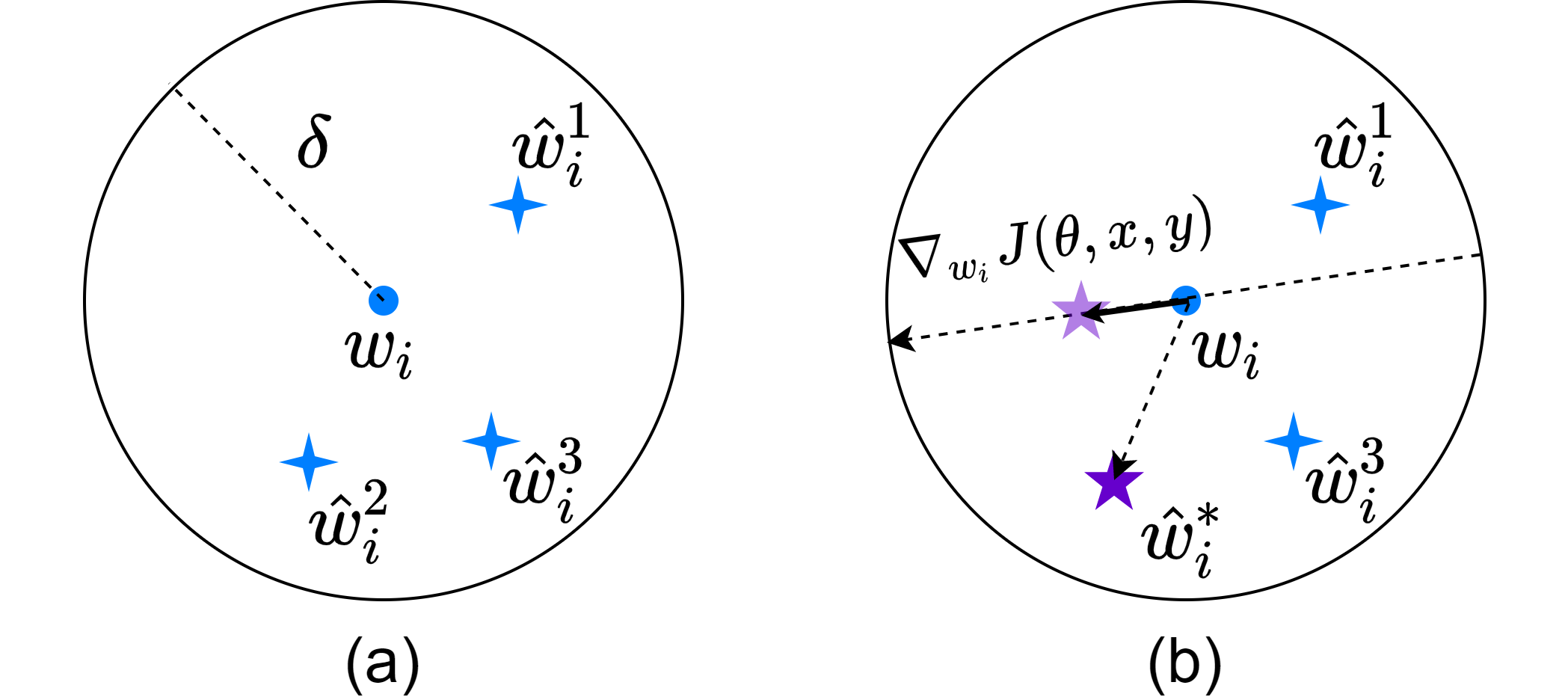}
    \caption{Strategies to pick optimal synonym to substitute word $w_i$. (a) Pick synonym $\hat{w}_i^*$ that minimizes the classification confidence among all the synonyms $\hat{w}_i^j \in S(w_i, \delta)$. (b) Pick synonym $\hat{w}_i^*$ that maximizes the product of the magnitude of gradient and the projected distance between $\hat{w}_i^*$ and $w_i$ in the gradient direction.}
    \label{fig:perturbation projection}
    \vspace{-1em}
\end{figure}

\subsection{Text Adversarial Examples}

Let $\mathcal{X}$ denote the input space containing all the possible input texts, $\mathcal{Y} = \{y_1, \cdots, y_m\}$ the output space and $\mathcal{D}$ the dictionary containing all the possible words in the input texts. Let $x=\langle w_1, \cdots, w_i, \cdots, w_n\rangle \in \mathcal{X}$ where $w_i \in \mathcal{D}$ denote an input sample consisting of $n$ words. A classifier $\phi$ is expected to learn a mapping $\mathcal{X} \to \mathcal{Y}$ so that for any sample $x$, the predicted label $\phi(x)$ equals its true label $y$ with high probability. Let $F(x,y)$ denote the logit output of  classifier $\phi$ on category $y$. The adversary adds an imperceptible perturbation $\Delta x$ on $x$ to craft an adversarial example $x_{adv}$ that misleads classifier $\phi$:
\begin{equation*}
    \begin{split}
        \phi(x_{adv})\neq \phi(x)=y, \  x_{adv} = x+\Delta x \quad
        s.t. \left\|\Delta x\right\|_p \leq \epsilon,
    \end{split}
\end{equation*}
where $\epsilon$ is a hyper-parameter for the perturbation upper bound, and $\left\| \cdot \right\|_p$ is the $L_p$-norm distance metric,
which often denotes the word substitution ratio $R(x, x_{adv})$ as the measure for the perturbation caused by synonym substitution:
\begin{equation*}
    R(x, x_{adv}) = \frac{1}{n} \sum_{i = 1}^n \mathbbm{1}_{w_i \neq w_i'}(w_i, w_i').
\end{equation*}
Here $\mathbbm{1}_{w_i \neq w_i'}$ is indicator function, 
$w_i \in x$ and $\ w_i' \in x_{adv}$.

\subsection{Generating Adversarial Examples}
\citet{mrkvsic2016counter} have shown that counter-fitting can help remove antonyms which are needlessly considered as ``similar words" in the original GloVe vector space to improve the capability of indicating semantic similarity. Thus, we post-process the GloVe vectors by counter-fitting and define a synonym set for each word $w_i \in x$ in the embedding space as follows:
\begin{equation}
    S(w_i, \delta)=\{\hat{w}_i \in \mathcal{D}~|~ \| \hat{w}_i- w_i\|_2 \leq \delta\}, 
    \label{eq:synonymSet}
\end{equation}
where $\delta$ is a hyper-parameter that constrains the maximum Euclidean distance for synonyms in the embedding space and we set $\delta = 0.5$ as in \citet{wang:IGA}.

Once we have the synonym set $S(w_i, \delta)$ for each word $w_i$, the next steps are for the optimal synonym selection and substitution order determination. 

\begin{algorithm}[tb]
    \algnewcommand\algorithmicinput{\textbf{Input:}}
    \algnewcommand\Input{\item[\algorithmicinput]}
    \algnewcommand\algorithmicoutput{\textbf{Output:}}
    \algnewcommand\Output{\item[\algorithmicoutput]}

    \caption{The FGPM Algorithm}
    \label{alg:FGPM}
	\begin{algorithmic}[1]
		\Input Benign sample $x=\langle w_1,\ \!\cdots\!,\!\  w_i,\ \!\cdots\!,\!\ w_n\rangle$
		\Input True label $y$ for $x$
		\Input Target classifier $\phi$
        \Input Upper bound distance for synonyms $\delta$
        \Input Maximum number of iterations $N$
        \Input Upper bound for word substitution ratio $\epsilon$
        \Output Adversarial example $x_{adv}$
		
		\State Initialize $x_{adv}^0=x$ 
		 \State Calculate $S(w_i, \delta)$ by Eq. \eqref{eq:synonymSet} for $w_i \in x_{adv}^{0}$
		\For{$k = 1 \to N$}
		    \State Construct candidate set $\mathcal{C}_s = \{\hat{w}_1^*,\ \!\cdots\!,\!\ \hat{w}_i^*,\ \!\cdots\!,\!\ \hat{w}_n^*\}$ by Eq. \eqref{eq:chooseWord}
		    \State Calculate optimal word $\hat{w}_*$ by Eq. \eqref{eq:replacement}
    		\State Substitute $w_* \in x_{adv}^{k-1}$ with $\hat{w}_*$ to obtain $x_{adv}^k$
    		\If{$\phi(x_{adv}^k) \neq y$ and $R(x_{adv}^k, x) < \epsilon$}
    		    \State \Return $x_{adv}^k$ \Comment{Succeed}
    		\EndIf
		\EndFor
		\State \Return None \Comment{Failed}
	\end{algorithmic} 
\end{algorithm}

\textbf{Word Substitution.} As shown in Figure~\ref{fig:perturbation projection} (a), for each word $w_i$, we expect to pick a word $\hat{w}_i^* \in S(w_i, \delta)$ that earns the most benefit to the overall substitution process of adversary generation, which we call optimal synonym. Due to the high complexity of finding optimal synonym, previous works \cite{kuleshov:GSA,wang:IGA} greedily pick a synonym $\hat{w}_i^* \in S(w_i, \delta)$ that minimizes the classification confidence:
\begin{equation*}
    \hat{w}_i^* = \mathop{\arg\max}_{\hat{w}_i^j \in S(w_i, \delta)} (F(x, y) - F(\hat{x}_i^j, y)),
\end{equation*}
where $\hat{x}_i^j=\langle w_1, \cdots, w_{i-1},\hat{w}_i^j, w_{i+1}, \cdots, w_n\rangle$. However, the selection process is time consuming as picking such a $\hat{w}_i^*$ needs $|S(w_i, \delta)|$ queries on the model. To reduce the calculation complexity, based on the local linearity of deep models, 
we use the product of gradient magnitude and projected distance between the original word and its synonym candidate in the gradient direction in the word embedding space to estimate the amount of change for the classification confidence. Specifically, as illustrated in Figure~\ref{fig:perturbation projection} (b), we first calculate the gradient $\nabla_{w_i}J(\theta, x, y)$ for each word $w_i$ where $J(\theta, x, y)$ is the loss function used for training. 
Then, we estimate the change by calculating $(\hat{w}_i^j - w_i) \cdot \nabla_{w_i}J(\theta, x, y)$.
To determine the optimal synonym $\hat{w}_i^*$, we choose a synonym with the maximum product value:
\begin{equation}
    \hat{w}_i^* = \mathop{\arg\max}_{\hat{w}_i^j \in S(w_i, \delta)} (\hat{w}_i^j - w_i) \cdot \nabla_{w_i}J(\theta, x, y).
    \label{eq:chooseWord}
\end{equation}

\textbf{Substitution Order.} For each word $w_i$ in text $x=\langle w_1, \cdots, w_i, \cdots, w_n\rangle$, we use the above word substitution strategy to choose its optimal substitution synonym and obtain a candidate set $\mathcal{C}_s = \{\hat{w}_1^*, \cdots, \hat{w}_i^*, \cdots, \hat{w}_n^*$\}. Then, we need to determine which word in $x$ should be substituted. Similar to the word substitution strategy, we pick a word $\hat{w}_i^* \in \mathcal{C}_s$, that has the biggest product of the gradient magnitude and the perturbation value projected in the gradient direction, to substitute $w_i \in x$:
\begin{equation}
    \hat{w}_* = \mathop{\arg\max}_{\hat{w}_i^* \in \mathcal{C}_s} (\hat{w}_i^* - w_i) \cdot \nabla_{w_i}J(\theta, x, y).
    \label{eq:replacement}
\end{equation}

In summary, to generate an adversarial example, we adopt the above word replacement and substitution order strategies for synonym substitution iteratively till the classifier makes a wrong prediction. The overall FGPM algorithm is shown in Algorithm \ref{alg:FGPM}.

To avoid the semantic drift caused by multiple substitutions at the same position of the text, we construct a candidate synonym set for the original sentence ahead of synonym substitution process and constrain all the substitutions with word $w_i\in x$ to the set, as shown at line 2 of Algorithm \ref{alg:FGPM}. We also set the upper bound for word substitution ratio $\epsilon = 0.25$ in our experiments. Note that at each iteration, previous query-based adversarial attacks need $\sum_{i=1}^n |S(w_i, \delta)|$ times of model queries ~\cite{kuleshov:GSA, ren:PWWS}, while FGPM just calculates the gradient by back-propagation once, leading to much higher efficiency.

\section{Adversarial Training with FGPM}
For image classification, \citet{Goodfellow:explaining} first propose adversarial training using the following objective function:
\begin{equation*}
    \tilde{J}(\theta, x,y) = \alpha J(\theta, x, y) + (1-\alpha) J(\theta, x_{adv}, y).
\end{equation*}
In recent years, numerous variants of adversarial training~\cite{kannan2018adversarial,Zhang2019Theoretically,song2019improving,Ding2020MMA} have been proposed to further enhance the robustness of models.

For text classification, previous works \cite{alzantot2018generating,ren:PWWS} have shown that incorporating their attack methods into standard adversarial training can improve the model robustness. Nevertheless, the improvement is limited. We argue that adversarial training requires plenty of adversarial examples generated based on instant model parameters in the training stage for better robustness enhancement.
Due to the inefficiency of text adversary generation, existing text attack methods based on synonym substitution could not provide sufficient adversaries for adversarial training. With the high efficiency of FGPM, we propose a new text defense method called \textit{Adversarial Training with FGPM enhanced by Logit pairing (ATFL)} to effectively improve the model robustness for text classification.

Specifically, we modify the objective function as follows:
\begin{equation*}
\begin{split}
    \tilde{J}(\theta, x,y) = \alpha J(\theta, x, y) + (1-\alpha) J(\theta, x_{adv}, y) \\  + \lambda \|F(x, \cdot) - F(x_{adv}, \cdot)\|.
\end{split}
\end{equation*}
where $x_{adv}$ is the adversarial example of each $x$ generated by FGPM based on the instant model parameters $\theta$ during training. In all our experiments, we set $\alpha = 0.5$ and $\lambda = 0.5$, and provide ablation study for $\alpha$ and $\lambda$ in Appendix.
As in \citet{kannan2018adversarial}, we train the model on adversarial examples and treat the logit similarity of benign examples and their adversarial counterparts as an regularizer to improve the model robustness rather than just adding a portion of adversarial examples of the already trained model into the training set and retrain the model.

\section{Experimental Results}
\label{sec:Experiments}

We evaluate FGPM with four attack baselines, and ATFL with two defense baselines, IBP and SEM, on three popular benchmark datasets involving CNN and RNN models. Code is available at \url{https://github.com/JHL-HUST/FGPM}.


\subsection{Experimental Setup}
We first introduce the experimental setup, including baselines, datasets and models used in experiments.

\textbf{Baselines.} For fair comparison, we restrict the perturbations of \citet{papernot2016crafting} within synonyms and denote this baseline as \textit{Papernot'}. To evaluate the attack effectiveness of FGPM, we compare it with four adversarial attacks, \textit{Papernot'}, GSA \cite{kuleshov:GSA}, PWWS \cite{ren:PWWS}, and IGA \cite{wang:IGA}. 
Furthermore, to validate the defense performance of our ATFL,
we take two competitive text defense methods, SEM~\cite{wang:IGA} and IBP~\cite{jia2019certified}, against the above word-level attacks. Due to the low efficiency of attack baselines, we randomly sample 200 examples on each dataset, and generate adversarial examples on various models. 

\textbf{Datasets.} We compare the proposed methods with baselines on three widely used benchmark datasets including \textit{AG's News}, \textit{DBPedia ontology} and \textit{Yahoo! Answers}~\cite{Zhang2015Dataset}. \textit{\textit{AG's News}} consists of news articles pertaining four classes:  World, Sports, Business and Sci/Tech. Each class includes 30,000 training examples and 1,900 testing examples. \textit{DBPedia ontology} is constructed by picking 14 non-overlapping classes from \textit{DBPedia 2014}, which is a crowd-sourced community effort to extract structured information from Wikipedia. For each of the 14 ontology classes, there are 40,000 training samples and 5,000 testing samples. \textit{Yahoo! Answers} is a topic classification dataset with 10 classes, and each class contains 140,000 training samples and 5,000 testing samples.

\textbf{Models.} We adopt several deep learning models that can achieve state-of-the-art performance on text classification tasks, including Convolution Neural Networks (CNNs) and Recurrent Neural Networks (RNNs). The embedding dimension of all models is 300 \cite{Mikolov2013Efficient}. Specifically, we replicate a CNN model from \citet{kim2014convolutional}, which consists of three convolutional layers with filter size of 5, 4, and 3 respectively, a dropout layer and a final fully connected layer. We also use a Long Short-Term Memory (LSTM) model which replaces the three convolutional layers of the CNN with three LSTM layers, each with 128 cells \cite{liu2016recurrent}. Lastly, we implement a Bi-directional Long Short-Term Memory (Bi-LSTM) model that replaces the three LSTM layers of the LSTM with a bi-directional LSTM layer having 128 forward direction cells and 128 backward direction cells.

\begin{table*}[tb]
\centering
\scalebox{0.9}{
\begin{tabular}{l|ccc|ccc|ccc}  
\toprule
\multirow{2}*{}  & \multicolumn{3}{c|}{\textit{AG's News}} & \multicolumn{3}{c|}{\textit{DBPedia}} & \multicolumn{3}{c}{\textit{Yahoo! Answers}} \\
\cmidrule(lr){2-4} \cmidrule(lr){5-7} \cmidrule(lr){8-10}
~ & { CNN} & { LSTM} & { Bi-LSTM} & { CNN} & { LSTM} & { Bi-LSTM} & { CNN} & { LSTM} & { Bi-LSTM} \\
\midrule
No Attack$^\dag$          & 92.3  & 92.6  & 92.5  & 98.7  & 98.8  & 99.0 & 72.3  & 75.1  & 74.9 \\
No Attack                 & 87.5  & 90.5  & 88.5  & 99.5  & 99.0  & 99.0 & 71.5  & 72.5  & 73.5 \\
\textit{Papernot'}  & 72.0  & 61.5  & 65.0  & 80.5  & 77.0  & 83.5 & 38.0  & 43.0  & 36.5 \\
GSA           & 45.5  & 35.0  & 40.0  & 52.0  & 49.0  & 53.5 & 21.5  & 19.5  & 19.0 \\
PWWS          & \underline{37.5}  & \underline{30.0}  & \underline{29.0}  & 55.5  & 52.5  & 50.0 & ~~\underline{5.5}  & \underline{12.5}  & 11.0 \\
IGA           & \textbf{30.0}  & \textbf{26.5}  & \textbf{25.5}  & \textbf{36.5}  & \textbf{38.5}  & \textbf{37.0} & ~~\textbf{3.5}  & ~~\textbf{5.5}  & ~~\textbf{7.0} \\
FGPM          & \underline{37.5}  & 31.0  & 32.0  & \underline{40.0}  & \underline{45.5}  & \underline{47.5} & ~~6.0  & 17.0  & \underline{10.5} \\
\bottomrule
\end{tabular}
}
\caption{The classification accuracy ($\%$) of different models under various competitive adversarial attacks. The first two rows of \textit{No Attack$^\dag$} and \textit{No Attack} show the model accuracy on the entire original test set and the sampled examples respectively. The \textbf{lowest classification accuracy} among the attacks is highlighted in \textbf{bold} to indicate the best attack effectiveness. The \underline{second lowest classification accuracy} is highlighted in \underline{underline}.}
\label{tab:Attack}
\end{table*}
\begin{table*}[tb]
    \centering
    \scalebox{0.9}{
    \begin{tabular}{l|ccc|ccc|ccc}
        \toprule
        ~ & { CNN} & { LSTM} & { Bi-LSTM} & { CNN} & { LSTM} & { Bi-LSTM} & { CNN} & { LSTM} & { Bi-LSTM} \\
        \midrule
        \textit{Papernot'} & 72.0* & 80.5 & 82.5 & 83.5 & 61.5* & 78.5 & 79.5 & 74.5 & 65.0* \\
        GSA                      & 45.5* & 80.0 & 80.0 & 84.5 & 35.0* & 73.0 & 81.5 & 72.5 & 40.0* \\
        PWWS                     & 37.5* & \textbf{70.5} & \textbf{70.0} & \underline{83.0} & 30.0* & \textbf{67.5} & 80.0 & \textbf{67.5} & 29.0* \\
        IGA                      & 30.0* & 74.5 & \underline{74.5} & 84.0 & 26.5* & \underline{71.5} & \underline{79.0} & \underline{71.0} & 25.5* \\
        FGPM                     & 37.5* & \underline{72.5} & \underline{74.5} & \textbf{81.0} & 31.0* & 73.5 & \textbf{77.5} & \textbf{67.5} & 32.0* \\
        \bottomrule
    \end{tabular}
    }
    \caption{The classification accuracy (\%) of different models for adversarial examples generated on other models on \textit{AG's News} for the transferability evaluation. * indicates that the adversarial examples are generated based on this model.}
    \label{tab:attack_transfer}
\vspace{-1em}
\end{table*}

\subsection{Evaluation on Attack Effectiveness}
To evaluate the attack effectiveness, we compare FGPM with the  baseline attacks in two aspects, namely model classification accuracy under attacks and transferability. 

\textbf{Classification Accuracy under Attacks.} In Table \ref{tab:Attack}, we provide the classification accuracy under FGPM and the competitive baseline attacks on three standard datasets. The more effective the attack method is, the more the classification accuracy the target model drops. We observe that IGA, adopting the genetic algorithm, can always achieve the best attack performance among all attacks. Compared with other attacks, FGPM could either achieve the best attack performance or on par with the best one. Especially, \textit{Papernot'}, the only gradient-based attack among the baselines, is inferior to FGPM
, indicating that the proposed gradient projection technique significantly improves the effectiveness of white-box word-level attacks. Besides, we also display some adversarial examples generated by FGPM in Appendix.

\textbf{Transferability.} The transferability of adversarial attack refers to the ability to reduce the classification accuracy of different models with adversarial examples generated on a specific model \cite{Goodfellow:explaining}, which is another serious threat in real-world applications. To illustrate the transferability of FGPM, we generate adversarial examples on each model by different attack methods and evaluate the classification accuracy of other models on these adversarial examples. Here, we evaluate the transferability of different attacks on \textit{AG's News}. 
As depicted in Table \ref{tab:attack_transfer}, the adversarial examples crafted by FGPM is on par with the best transferability performance among the baselines.

\begin{table*}[t]
\centering
\scalebox{0.9}{
\begin{tabular}{l|rrr|rrr|rrr}  
\toprule
\multirow{2}*{}  & \multicolumn{3}{c|}{\textit{AG's News}} & \multicolumn{3}{c|}{\textit{DBPedia}} & \multicolumn{3}{c}{\textit{Yahoo! Answers}} \\
\cmidrule(lr){2-4} \cmidrule(lr){5-7} \cmidrule(lr){8-10}
~ & { CNN} & { LSTM} & { Bi-LSTM} & { CNN} & { LSTM} & { Bi-LSTM} & { CNN} & { LSTM} & { Bi-LSTM} \\
\midrule
\textit{Papernot'}          & 74  & 1,676  & 4,401  & 145  & 2,119  & 6,011 & 120  & 9,719  & 19,211 \\
GSA           & 276  & 643  & 713  & 616  & 1,006  & 1,173 & 1,257  & 2,234  & 2,440 \\
PWWS          & 122  & 28,203  & 28,298  & 204  & 34,753  & 35,388 & 643  & 98,141  & 100,314 \\
IGA           & 965  & 47,142  & 91,331  & 1,369  & 69,770  & 74,376 & 893  & 132,044  & 123,976 \\
FGPM    & \textbf{8}    & \textbf{29}    & \textbf{29}  & \textbf{8}  & \textbf{34}  & \textbf{33} & \textbf{26}  & \textbf{193}  &  \textbf{199} \\
\bottomrule
\end{tabular}
}
\caption{Comparison on the total running time (in seconds) for generating $200$ adversarial instances.}
\label{tab:attack_efficiency}
\end{table*}

\begin{table*}[tbp]
\centering
\scalebox{0.9}{
\begin{tabular}{l|l|rrrc|rrrc|rrrc}  
\toprule
\multirow{2}*{Dataset} & \multirow{2}*{Attack} & \multicolumn{4}{c|}{CNN} & \multicolumn{4}{c|}{LSTM} & \multicolumn{4}{c}{Bi-LSTM} \\
\cmidrule(lr){3-6} \cmidrule(lr){7-10} \cmidrule(lr){11-14}
~ & ~ & NT & SEM & IBP & ATFL & NT & SEM & IBP & ATFL & NT & SEM & IBP & ATFL \\
\midrule
\multirow{7}*{\textit{\shortstack{AG's\\News}}} 
  & No Attack$^\dag$ & \textbf{92.3}   & 89.7  & 89.4 & 91.8   & \textbf{92.6} & 90.9  & 86.3 & 92.0  & \textbf{92.5}  & 91.4 & 89.1 & 92.1  \\
~ & No Attack  & 87.5    & 87.5  & 87.5 & \textbf{89.0}        & 90.5  & 90.5 & 84.5  & \textbf{91.5}   & 88.5   & \textbf{91.0} & 87.0  & 89.5     \\
~ & \textit{Papernot'} & 72.0  & 84.5  & 87.5   & \textbf{88.0}  & 61.5   & 89.5 & 81.5  & \textbf{90.0}   & 65.0   & \textbf{90.0}  & 86.0 & 89.0  \\
~ & GSA           & 45.5     & 80.0  & 86.0 & \textbf{88.0}    & 35.0  & 85.5  & 79.5 & \textbf{88.0}   & 40.0   & \textbf{87.5} & 79.0  & \textbf{87.5}  \\
~ & PWWS          & 37.5     & 80.5  & 86.0 & \textbf{88.0}    & 30.0  & 86.5  & 79.5 & \textbf{88.0}   & 29.0   & \textbf{87.5} & 75.5  & \textbf{87.5}  \\
~ & IGA           & 30.0     & 80.0 & 86.0 & \textbf{88.0}     & 26.5  & 85.5 & 79.5  & \textbf{88.0}   & 25.5   & \textbf{87.5} & 79.0  & \textbf{87.5}     \\
~ & FGPM    & 37.5     & 78.5  & 86.5 & \textbf{88.0}    & 31.0  & 85.5 & 80.0  & \textbf{88.0}   & 32.0   & 84.5 & 80.0  & \textbf{87.5}  \\

\midrule
\multirow{7}*{\textit{DBPedia}}
  & No Attack$^\dag$    & \textbf{98.7}     & 98.1  & 97.4  & 98.4  & \textbf{98.8} & 98.5  & 93.1 & 98.7  & \textbf{99.0} & 98.7 & 94.7 & 98.6  \\
~ & No Attack     & \textbf{99.5}     & 97.5  & 97.0 & 98.0  & 99.0  & \textbf{99.5} & 95.0 & \textbf{99.5}  & \textbf{99.0}  & 98.0 & 94.5 & \textbf{99.0}  \\
~ & \textit{Papernot'} & 80.5  & 97.0 & 97.0 & \textbf{98.0}   & 77.0  & \textbf{99.5} & 91.0 & \textbf{99.5}   & 83.5  & 98.0 & 92.5 & \textbf{99.0}  \\
~ & GSA           & 52.0     & 96.0  & 97.0 & \textbf{98.0}   & 49.0  & \textbf{99.0} & 84.5 & 98.5   & 53.5  & 98.0 & 89.5 & \textbf{99.0}  \\
~ & PWWS          & 55.5     & 95.5  & 97.0 & \textbf{98.0}   & 52.5  & \textbf{99.5} & 84.0 & 98.5  & 50.0  & 95.0  & 89.5 & \textbf{99.0}  \\
~ & IGA           & 36.5     & 95.5  & 97.0 & \textbf{98.0}   & 38.5  & \textbf{99.0} & 84.5 & 98.0   & 37.0  & 97.0 & 90.0  & \textbf{99.0}  \\
~ & FGPM    & 40.0     & 94.0  & 97.0 & \textbf{98.0}   & 45.5  & \textbf{99.0} & 85.0 & 98.5   & 47.5  & 98.0 & 89.5 & \textbf{99.0}  \\

\midrule
\multirow{7}*{\textit{\shortstack{Yahoo!\\Answers}}}
  & No Attack$^\dag$    & \textbf{72.3}    & 70.0 & 64.2 & 71.0   & \textbf{75.1}     & 72.8 & 51.2 & 74.2  & \textbf{74.9}    & 72.9     & 59.0 & 74.3  \\
~ & No Attack     & 71.5     & 67.0 & 64.5 & \textbf{72.0}     & 72.5     & 69.5  & 50.5 & \textbf{74.0}     & \textbf{73.5} & 69.5  & 56.0 & 72.0  \\
~ & \textit{Papernot'} & 38.0   & 64.0 & 63.5 & \textbf{69.0}   & 43.0     & 67.0 & 41.0 & \textbf{71.0}  & 36.5   & 66.5     & 53.0  & \textbf{70.5}  \\
~ & GSA           & 21.5     & 59.5   & 61.0 & \textbf{63.0}    & 19.5     & 63.0 & 30.0 & \textbf{69.5}  & 19.0    & 62.5          & 39.5 & \textbf{64.5}  \\
~ & PWWS          & 5.5      & 59.0   & 61.0 & \textbf{62.5}    & 12.5     & 63.0 & 30.0 & \textbf{68.5}  & 11.0      & 62.5 & 40.0 & \textbf{65.5}  \\
~ & IGA           & 3.5      & 59.0   & 61.0 & \textbf{62.5}    & 5.5      & 62.5 & 31.5 & \textbf{67.5}     & 7.0 & 62.0   & 40.5 & \textbf{64.0}    \\
~ & FGPM    & 6.0      & 61.0   & 63.0 & \textbf{64.0}    & 17.0     & 63.0 & 35.0 & \textbf{68.5}  & 10.5      & \textbf{64.5} & 41.5 & 63.5  \\

\bottomrule
\end{tabular}
}
\caption{The classification accuracy ($\%$) of three competitive defense methods under various adversarial attacks on the same set of 200 randomly selected samples for the three standard datasets.}
\label{tab:Defence Effect}
\vspace{-0.5em}
\end{table*}

\subsection{Evaluation on Attack Efficiency}
The attack efficiency is important for evaluating attack methods, especially if we would like to incorporate the attacks into adversarial training as a defense method. Adversarial training needs highly efficient adversary generation so as to effectively promote the model robustness. Thus, we evaluate the total time (in seconds) of generating $200$ adversarial examples on the three datasets by various attacks. As shown in Table \ref{tab:attack_efficiency}, the average time of generating $200$ adversarial examples by FGPM is nearly $20$ times faster than GSA, the second fastest synonym substitution based attack but with weaker attack performance and lower transferability than FGPM. Moreover, FGPM is on average $970$ times faster than IGA, which produces the maximum degradation of the classification accuracy among the baselines. Though \textit{Papernot'} crafts adversarial examples based on gradient, which makes each iteration faster, it needs much more iterations to obtain adversarial examples due to low attack effectiveness. On average, FGPM is about 78 times faster than \textit{Papernot'}.


\subsection{Evaluation on Adversarial Training}
From the above analysis, we see that compared with the competitive attack baselines, FGPM can achieve much higher efficiency with good attack performance and transferability. Such performance enables us to implement effective adversarial training and scale to large neural networks and datasets.
In this subsection, we evaluate the defence performance of ATFL and conduct comparison with SEM and IBP against adversarial examples generated by the above attacks. Here we focus on two factors, defense against adversarial attacks and defense against transferability.

\textbf{Defense against Adversarial Attacks.} We use the above attacks on models trained by various defense methods to evaluate the defense performance. The results are shown in Table \ref{tab:Defence Effect}. For normal training (NT), the classification accuracy on all datasets drops dramatically under different adversarial attacks. In contrast, both SEM and ATFL can promote the model robustness stably and effectively among all models and datasets. 
IBP, originally proposed for CNN to defend the adversarial attacks in image domain, can improve the robustness of CNN on three datasets but with much higher computation cost. More importantly, with many restrictions added on the architectures, the model hardly converges when trained on LSTM and Bi-LSTM, resulting in both weakened generalization and adversarial robustness instead. Compared with SEM, moreover, ATFL can obtain higher classification accuracy on benign data,
and is very competitive under almost all adversarial attacks. 

\begin{table*}[!ht]
\centering
\scalebox{0.9}{
\begin{tabular}{l|cccc|cccc|cccc}  
\toprule
\multirow{2}*{Attack} & \multicolumn{4}{c|}{CNN} & \multicolumn{4}{c|}{LSTM} & \multicolumn{4}{c}{Bi-LSTM} \\
\cmidrule(lr){2-5} \cmidrule(lr){6-9} \cmidrule(lr){10-13}
~ & NT & SEM & IBP & ATFL & NT & SEM & IBP & ATFL & NT & SEM & IBP & ATFL \\
\midrule
\textit{Papernot'} & 72.0* & 87.0 & 87.0 & \textbf{88.5}   & 80.5   & 91.0 & 82.0 & \textbf{92.0}     & 82.5    & \textbf{91.0} & 86.0 & 90.0  \\
GSA                  & 45.5*    & 87.0  & 87.0 & \textbf{88.5}   & 80.0   & 90.5 & 83.0 & \textbf{91.0}     & 80.0    & \textbf{91.0} & 87.5 & 90.0  \\
PWWS                 & 37.5*    & 87.0  & 87.0 & \textbf{88.5}   & 70.5   & \textbf{90.5} & 83.0 & \textbf{90.5}     & 70.0    & \textbf{90.5} & 86.5 & 90.0  \\
IGA                  & 30.0*    & 87.0  & 87.0 & \textbf{88.5}   & 74.5   & 90.5 & 83.5 & \textbf{91.0}     & 74.5    & \textbf{90.5} & 86.5 & 89.5  \\
FGPM                 & 37.5*    & 87.0  & 87.5 & \textbf{88.5}   & 72.5   & 90.5 & 83.0 & \textbf{91.5}     & 74.5    & \textbf{91.0} & 86.5 & 90.0  \\

\midrule
\textit{Papernot'} & 83.5 & 87.5 & 87.5 & \textbf{88.0}   & 61.5*   & \textbf{91.0} & 82.0 & \textbf{91.0}     & 78.5    & \textbf{91.0} & 86.5 & 89.5  \\
GSA                   & 84.5    & 87.0  & 87.5 & \textbf{88.5}  & 35.0*   & 90.5 & 83.5 & \textbf{91.0}     & 73.0    & \textbf{91.0} & 86.5 & 89.5  \\
PWWS                  & 83.0    & 87.0  & 87.5 & \textbf{89.0}  & 30.0*   & \textbf{90.5} & 85.0 & \textbf{90.5}     & 67.5    & \textbf{90.5} & 86.5 & 90.0  \\
IGA                   & 84.0    & 87.0  & 87.5 & \textbf{88.5}  & 26.5*   & 90.5 & 83.5 & \textbf{91.5}     & 71.5    & \textbf{91.0} & 87.0 & 90.0  \\
FGPM                  & 81.0    & 87.5  & 87.5 & \textbf{89.0}  & 31.0*   & 90.5 & 83.5 & \textbf{91.5}     & 73.5    & \textbf{91.0} & 87.0 & 89.5  \\

\midrule
\textit{Papernot'} & 79.5 & 88.0  & 87.0 & \textbf{88.5}  & 74.5   & \textbf{91.0} & 82.5 & \textbf{91.0}     & 65.0*    & \textbf{91.0} & 86.5 & 89.0  \\
GSA                   & 81.5    & 87.0  & 87.5 & \textbf{88.5}  & 72.5   & 90.5 & 84.0 & \textbf{91.0}     & 40.0*    & \textbf{91.0} & 87.5 & 90.0  \\
PWWS                  & 80.0    & 86.5  & 87.0 & \textbf{89.0}  & 67.5   & 90.5 & 83.5 & \textbf{91.5}     & 29.0*    & \textbf{90.5} & 87.0 & 90.0  \\
IGA                   & 79.0    & 87.0  & 87.0 & \textbf{88.5}  & 71.0   & 90.5 & 83.5 & \textbf{91.0}     & 25.5*    & \textbf{91.0} & 86.5 & 89.5  \\
FGPM                  & 77.5    & 87.5  & 87.5 & \textbf{89.0}  & 67.5   & 90.5 & 83.5 & \textbf{91.0}     & 32.0*    & \textbf{91.0} & 87.0 & 89.5  \\

\bottomrule
\end{tabular}
}
\caption{The classification accuracy ($\%$) of various models under competitive defenses for adversarial examples generated on other models on \textit{AG's News} for evaluating the defense performance against transferability. * indicates the white-box attacks.}
\label{tab:Transferability}

\vspace{1.5em}

\centering
\scalebox{0.9}{
\begin{tabular}{l|l|p{1.2cm}<{\centering} p{1.2cm}<{\centering} p{1.2cm}<{\centering} p{1.2cm}<{\centering} p{1.2cm}<{\centering} p{1.2cm}<{\centering} p{1.2cm}<{\centering}}  
\toprule
Model                     & Attack & NT & Standard & TRADES & MMA & MART & CLP & ALP \\
\midrule
\multirow{7}*{CNN}        & No Attack$^\dag$ & \textbf{92.3} & \textbf{92.3} & 92.1 & 91.1 & 91.2 & 91.7  & 91.8  \\
~                         & No Attack  & 87.5 & 89.5 & 89.5 & 87.5 & 87.0 & \textbf{90.5}  & 89.0  \\
~          & \textit{Papernot'}  & 72.0 & 85.5 & 67.0 & 83.5 & 83.5 & 73.0  & \textbf{88.0}  \\
~                         & GSA        & 45.5 & 77.5 & 36.5 & 69.0 & 73.0 & 42.5  & \textbf{88.0}  \\
~                         & PWWS       & 37.5 & 77.0 & 33.5 & 70.5 & 73.0 & 38.5  & \textbf{88.0}  \\
~                         & IGA        & 30.0 & 75.0 & 29.0 & 67.5 & 72.0 & 30.0  & \textbf{88.0}  \\
~                         & FGPM       & 37.5 & 78.0 & 40.0 & 73.5 & 74.5 & 38.5  & \textbf{88.0}  \\

\midrule
\multirow{7}*{LSTM}       & No Attack$^\dag$ & \textbf{92.6} & \textbf{92.6} & 91.9 & 91.3 & 90.8 & 92.1  & 92.0  \\
~                         & No Attack  & 90.5 & \textbf{92.0} & 90.5 & 89.0 & 90.0 & 91.0  & 91.5  \\
~          & \textit{Papernot'}  & 61.5 & 88.0 & 66.0 & 86.0 & 86.0 & 69.0  & \textbf{90.0}  \\
~                         & GSA        & 35.0 & 83.0 & 37.5 & 78.0 & 79.0 & 40.5  & \textbf{88.0}  \\
~                         & PWWS       & 30.0 & 84.0 & 32.0 & 78.0 & 79.5 & 46.5  & \textbf{88.0}  \\
~                         & IGA        & 26.5 & 83.0 & 24.0 & 77.5 & 79.5 & 34.0  & \textbf{88.0}  \\
~                         & FGPM       & 31.0 & 83.0 & 32.5 & 81.5 & 80.5 & 41.0  & \textbf{88.0}  \\

\midrule
\multirow{7}*{Bi-LSTM}    & No Attack$^\dag$ & 92.5 & \textbf{92.8} & 92.4 & 91.4 & 92.3 & 92.4  & 92.1  \\
~                         & No Attack  & 88.5 & 89.5 & \textbf{90.5} & 88.5 & 90.0 & \textbf{90.5}  & 89.5  \\
~          & \textit{Papernot'}  & 65.0 & \textbf{89.5} & 65.5 & 85.5 & 86.0  & 66.0 & 89.0  \\
~                         & GSA        & 40.0 & 86.0 & 35.5 & 81.0 & 80.5  & 38.5  & \textbf{87.5}  \\
~                         & PWWS       & 29.0 & 86.5 & 30.0 & 80.0 & 80.5  & 52.0  & \textbf{87.5}  \\
~                         & IGA        & 25.5 & 86.0 & 29.0 & 78.5 & 80.0  & 34.5  & \textbf{87.5}  \\
~                         & FGPM       & 32.0 & 86.5 & 32.0 & 82.0 & 80.5  & 46.0  & \textbf{87.5}  \\

\bottomrule
\end{tabular}
}
\caption{The classification accuracy ($\%$) of different classification models adversarially trained with different regularization under various adversarial attacks on the same set of 200 randomly selected samples for the \textit{AG's News} dataset.}
\label{tab:DefenseRegularization}
\vspace{-0.5em}
\end{table*}
\textbf{Defense against Transferability.} As transferability poses a serious concern in real-world applications, a good defense method should not only defend the adversarial attack but also resist the transferability of adversarial examples. To evaluate the ability of blocking transferability, we evaluate each model's classification accuracy on adversarial examples generated by different attack methods under normal training on \textit{AG's News}. As shown in Table \ref{tab:Transferability}, ATFL is much more successful in blocking the transferability of adversarial examples than the defense baselines on CNN and LSTM and achieve similar accuracy to SEM on Bi-LSTM.

In summary, ATFL can significantly promote the model robustness, block the transferability of adversarial examples successfully and achieve better generalization on benign data compared with other defenses.
Moreover, when applied to complex models and large datasets, ATFL maintains stable and effective performance.

\subsection{Evaluation on Adversarial Training Variants}
\label{sec:VAT}
Many variants of adversarial training, such as CLP and ALP \cite{kannan2018adversarial}, TRADES \cite{Zhang2019Theoretically}, MMA \cite{Ding2020MMA}, MART \cite{Wang2020Improving}, have tried to adopt different regularizations to improve the effectiveness of adversarial training for image data. The loss functions for these variants are depicted in Appendix. Here we try to answer the following question: can these variants also bring improvement for texts?

To validate the effectiveness of these variants, we run the above methods on \textit{AG's News} with three models. As shown in Table \ref{tab:DefenseRegularization}, standard adversarial training can improve both generalization and robustness of the models. Among the variants, however, only ALP can further improve the performance of adversarial training. Some recent variants (e.g. TRADES, CLP) that work very well for images significantly degrade the performance of standard adversarial training for texts, indicating that we need more specialized adversarial training methods for texts.

\section{Conclusion}
In this work, we propose an efficient gradient based synonym substitution adversarial attack method, called \textit{Fast Gradient Projection Method} (FGPM).
Empirical evaluations on three widely used benchmark datasets demonstrate that FGPM is about 20 times faster than the current fastest synonym substitution based adversarial attack method, and FGPM can achieve similar attack performance and transferability. 
With such high efficiency, we introduce an effective defense method called \textit{Adversarial Training with FGPM enhanced by Logit pairing} (ATFL) for text classification. 
Extensive experiments demonstrate that ATFL can significantly promote the model robustness, block the transferability of adversarial examples effectively, and achieve better generalization on benign data than text defense baselines. Besides, we find that recent successful regularizations of adversarial training for image data actually degrade the performance of adversarial training in text domain, suggesting the need for more specialized adversarial training methods for text data. 

Our work offers a way to adopt gradient for adversarial attack in discrete space, making it possible to adapt successful gradient based image attacks for text adversarial attacks. Besides, considering the prosperity of adversarial training for image data and high efficiency of gradient based methods, we hope our work could inspire more research of adversarial training in text domain.


\section*{Acknowledgement}

This work is supported by National Natural Science Foundation (62076105) and Microsft Research Asia Collaborative Research Fund (99245180). We thank Kai-Wei Chang for helpful suggestions on our work.

\bibliography{aaai21.bib}

\newpage
\newpage

\appendix
\section{Appendix}

\label{sec_appendix}

\subsection{Ablation study}
In our experiments, we adopt $\alpha=0.5$ and $\lambda=0.5$ without any parameter tuning as used by the FGSM adversarial training \cite{Goodfellow:explaining}. To further explore the impact of hyper-parameters $\alpha$ and $\beta$, we try various values of parameter $\alpha$ using $0.1, 0.3, 0.5, 0.7, 0.9$ while $\lambda$ is fixed to $0.5$ for our CNN model on \textit{AG’s News} and then do similar experiments for $\lambda$. For parameter $\alpha$, as shown in Table \ref{tab:ablation_alpha}, the clean accuracy is the highest at 91.9\% when $\alpha=0.3$ or $0.9$. For the defense efficacy, the models achieve the highest accuracy of 88\% when $\alpha=0.5$, but the lowest accuracy under various attacks is still good at 85.5\%. For parameter $\lambda$, as shown in Table \ref{tab:ablation_beta}, the clean accuracy is the highest at 92\% when $\lambda=0.7$. The accuracies for $\lambda=0.5$ and $0.9$ are roughly the same against various attacks, which achieve the best robustness of 88\%. The lowest accuracy for various $\lambda$ is still good at 85.5\%. Thus, the parameters have little impact on the defense efficacy.

\begin{table}[!ht]
\centering
\begin{tabular}{l|ccccc}  
\toprule
\diagbox{Attack}{$\alpha$} & 0.1 & 0.3 & 0.5 & 0.7 & 0.9 \\
\midrule
No Attack$^\dag$          & 91.6 & 91.9&91.8 &91.6 &91.9\\
No Attack                 & 88.5 & 87.5&89.0 &89.0 &90.0\\
PWWS          & 86.5 & 85.5&88.0 &86.5 &86.0\\
IGA           & 86.5 & 86.5&88.0 &86.0 &86.0\\
FGPM          &  86.5 & 86.5&88.0 &86.5 &86.5\\
\bottomrule
\end{tabular}
\caption{The classification accuracy (\%) of ATFL using various values of $\alpha$ where $\lambda$ is fixed to $0.5$ on the CNN model on \textit{AG's News} under various adversarial attacks.}
\label{tab:ablation_alpha}

\vspace{1em}
\begin{tabular}{l|ccccc}  
\toprule
\diagbox{Attack}{$\lambda$} & 0.1 & 0.3 & 0.5 & 0.7 & 0.9 \\
\midrule
No Attack$^\dag$          & 91.8 & 91.9&91.8 &92.0 &91.6\\
No Attack                 & 89.5 & 88.5&89.0 &87.0 &89.5\\
PWWS          & 85.5 & 86.0&88.0 &85.5 &87.0\\
IGA           & 85.5 & 86.0&88.0 &85.5 &87.0\\
FGPM          &  86.0 & 87.0&88.0 &85.5 &88.0\\
\bottomrule
\end{tabular}
\caption{The classification accuracy (\%) of ATFL using various values of $\lambda$ where $\alpha$ is fixed to $0.5$ on the CNN model on \textit{AG's News} under various adversarial attacks.}
\label{tab:ablation_beta}
\end{table}
\begin{table*}[!htb]
\centering
\begin{tabular}{c|c}  
\toprule
Defense Method & Loss Function \\
\midrule
Standard    & $\alpha CE(F(x,\cdot), y) + (1-\alpha) CE(F(x_{adv},\cdot), y)$\\
TRADES      & $CE(F(x,\cdot),y) + \lambda \cdot \|F(x,\cdot) - F(x_{adv},\cdot)\|$\\
MMA         & $CE(F(x,\cdot), y) \cdot \mathbbm{1}(\phi(x) \neq y) + CE(F(x_{adv},\cdot), y) \cdot \mathbbm{1}(\phi(x) = y)$\\
MART        & $BCE(F(x_{adv},\cdot), y) + \lambda \cdot KL(F(x, \cdot) \| F(x_{adv}, \cdot)) \cdot (1 - F(x, y))$\\
CLP         & $CE(F(x, \cdot), y) + \lambda \cdot \|F(x, \cdot) - F(x', \cdot)\|$\\
ALP        & $\alpha CE(F(x,\cdot), y) + (1-\alpha) CE(F(x_{adv},\cdot), y) + \lambda \cdot \|F(x, \cdot) - F(x_{adv}, \cdot)\|$\\
\bottomrule
\end{tabular}
\caption{The loss functions for different variations of adversarial training.}
\label{tab:loss_function}
\end{table*}

\subsection{Adversarial Training Variants}
\label{app:VAT}
In this subsection, we provide loss functions of different variations of adversarial training used in Section \ref{sec:VAT} in Table \ref{tab:loss_function}. We assume that the objective function $J(\theta, x, y)$ denotes the cross entropy between the logit output $F(x, \cdot)$ and true label $y$, namely $CE(F(x, \cdot), y)$. $x'$ in CLP is another randomly selected clean training example. And the boosted cross entropy (BCE) loss is defined as follows:
\begin{equation*}
    \resizebox{0.99\hsize}{!}{ $
    BCE(F(x_i,\cdot), y_i) = - \log(F(x_i, y_i)) - \log (1 - \mathop{\max}_{k \neq y_i} F(x_i, k)) $}
\end{equation*}

\begin{table*}[hbp]
\centering
\scalebox{0.9}{
\begin{tabular}{l|c|c}
\toprule
~ & Prediction & Text \\
\midrule
Original & \tabincell{c}{Business \& Finance \\ Confidence: 68.2\%} & \multicolumn{1}{m{13cm}}{My college age daughter and I want to move to Manhatten. We need to find housing and \emp{jobs}. Any suggestions? My daughter has done alittle modeling. I am an industrial engineer and would love to get back into the \emp{apparel} industry. My daughter would like to transfer to NYU as a film major. It is really really expensive. A one 2 bedroom apartment, if you can find one, will cost you at least \$ 3,000.} \\
\midrule
Adversarial & \tabincell{c}{Education \& Reference \\ Confidence: 44.3\%}  & \multicolumn{1}{m{13cm}}{My college age daughter and I want to move to Manhatten. We need to find housing and \emp{work}. Any suggestions? My daughter has done alittle modeling. I am an industrial engineer and would love to get back into the \emp{clothe} industry. My daughter would like to transfer to NYU as a film major. It is really really expensive. A one 2 bedroom apartment, if you can find one, will cost you at least \$ 3,000.} \\
\midrule
Original & \tabincell{c}{Education \& Reference \\ Confidence: 58.7\%} & \multicolumn{1}{m{13cm}}{I need more time management skills, can anyone help? I'm a high school \emp{student} who is also doing another course but I can't find enough time to get everthing done, I'm a week or two behind already. Please help! You can try managing your time by prioritizing by importance and / or deadline. You can start attending to requirements that are due at an earlier date. It also helps if you regularly study as compared to \emp{studying} (or craming) for a test the night before, because you are always prepared. It will also be a good idea to start right away when projects are given, as also not to cram. You'll feel less stressed. There is no best time management for everyone. Each one has his own time management technique, so try and find your own style that works best for you. I hope this helps! Good luck!} \\
\midrule
Adversarial & \tabincell{c}{Business \& Finance \\ Confidence: 57.3\%} & \multicolumn{1}{m{13cm}}{I need more time management skills, can anyone help? I'm a high school \emp{schoolchildren} who is also doing another course but I can't find enough time to get everthing done, I'm a week or two behind already. Please help! You can try managing your time by prioritizing by importance and / or deadline. You can start attending to requirements that are due at an earlier date. It also helps if you regularly study as compared to \emp{explore} (or craming) for a test the night before, because you are always prepared. It will also be a good idea to start right away when projects are given, as also not to cram. You'll feel less stressed. There is no best time management for everyone. Each one has his own time management technique, so try and find your own style that works best for you. I hope this helps! Good luck!} \\
\bottomrule
\end{tabular}}
\caption{The adversarial examples generated by FGPM on \textit{Yahoo! Answers} using CNN model.}
\label{tab:adversarial_examples_on_cnn}
\end{table*}

\newpage
\subsection{Adversarial Examples Generated by FGPM}
We randomly pick some adversarial examples generated by FGPM on \textit{Yahoo! Answers} using normally trained CNN, LSTM and Bi-LSTM models, and display the generated examples in Table~\ref{tab:adversarial_examples_on_cnn}, \ref{tab:adversarial_examples_on_lstm} and \ref{tab:adversarial_examples_on_bilstm}. We \underline{\textbf{\textit{underline and bold}}} the location of the perturbation for emphasis. It can be observed that the perturbation imposed by the 
proposed attack method on the three models is different. Note that the adversarial examples are basically consistent with the benign samples in terms of syntax and semantics.

\begin{table*}[hbp]
\centering
\scalebox{0.9}{
\begin{tabular}{l|c|c}
\toprule
& Prediction & Text \\
\midrule
Original & \tabincell{c}{Business \& Finance \\ Confidence: 76.2\%} & \multicolumn{1}{m{13cm}}{My \emp{college} age daughter and I want to move to Manhatten. We need to find housing and jobs. Any suggestions? My daughter has done alittle \emp{modeling}. I am an industrial engineer and would love to get back into the apparel industry. My daughter would like to transfer to NYU as a film major. It is really really expensive. A one 2 bedroom apartment, if you can find one, will cost you at least \$ 3,000.} \\
\midrule
Adversarial & \tabincell{c}{Education \& Reference \\ Confidence: 47.7\%} & \multicolumn{1}{m{13cm}}{My \emp{universities} age daughter and I want to move to Manhatten. We need to find housing and jobs. Any suggestions? My daughter has done alittle \emp{models}. I am an industrial engineer and would love to get back into the apparel industry. My daughter would like to transfer to NYU as a film major. It is really really expensive. A one 2 bedroom apartment, if you can find one, will cost you at least \$ 3,000.} \\
\midrule
Original & \tabincell{c}{Education \& Reference \\ Confidence: 81.6\%} &  \multicolumn{1}{m{13cm}}{I need more time \emp{management} skills, can anyone help? I'm a high school student who is also doing another course but I can't find enough time to get everthing done, I'm a week or two behind already. Please help! You can try managing your time by prioritizing by importance and / or deadline. You can start attending to requirements that are due at an earlier date. It also helps if you regularly study as compared to studying (or craming) for a test the night before, because you are always prepared. It will also be a good idea to start right away when projects are given, as also not to cram. You'll feel less stressed. There is no best time management for everyone. Each one has his own time management technique, so try and find your own style that works best for you. I hope this helps! Good luck!} \\
\midrule
Adversarial & \tabincell{c}{Business \& Finance \\ Confidence: 43.6\%} &  \multicolumn{1}{m{13cm}}{I need more time \emp{governance} skills, can anyone help? I'm a high school student who is also doing another course but I can't find enough time to get everthing done, I'm a week or two behind already. Please help! You can try managing your time by prioritizing by importance and / or deadline. You can start attending to requirements that are due at an earlier date. It also helps if you regularly study as compared to studying (or craming) for a test the night before, because you are always prepared. It will also be a good idea to start right away when projects are given, as also not to cram. You'll feel less stressed. There is no best time management for everyone. Each one has his own time management technique, so try and find your own style that works best for you. I hope this helps! Good luck!} \\
\bottomrule
\end{tabular}
}
\vspace{-0.3em}
\caption{The adversarial examples generated by FGPM on \textit{Yahoo! Answers} using LSTM model.}
\label{tab:adversarial_examples_on_lstm}
\end{table*}

\begin{table*}[hbp]
\centering
\scalebox{0.9}{
\begin{tabular}{l|c|c}
\toprule
& Prediction & Text \\
\midrule
Original & \tabincell{c}{Business \& Finance \\ Confidence: 63.6\%} & \multicolumn{1}{m{13cm}}{My college age daughter and I want to move to Manhatten. We need to find housing and \emp{jobs}. Any suggestions? My daughter has done alittle modeling. I am an industrial engineer and would love to get back into the \emp{apparel} industry. My daughter would like to transfer to NYU as a film major. It is really really expensive. A one 2 \emp{bedroom} apartment, if you can find one, will cost you at least \$ 3,000.} \\
\midrule
Adversarial & \tabincell{c}{Education \& Reference \\ Confidence: 50.9\%} & \multicolumn{1}{m{13cm}}{My college age daughter and I want to move to Manhatten. We need to find housing and \emp{work}. Any suggestions ? My daughter has done alittle modeling. I am an industrial engineer and would love to get back into the \emp{garment} industry. My daughter would like to transfer to NYU as a film major. It is really really expensive. A one 2 \emp{room} apartment, if you can find one, will cost you at least \$ 3,000.} \\
\midrule
Original & \tabincell{c}{Education \& Reference \\ Confidence: 88.3\%} & \multicolumn{1}{m{13cm}}{I need more time management \emp{skills}, can anyone help? I'm a high school student who is also \emp{doing} another course but I can't find enough time to get \emp{everthing} done, I'm a week or two behind already. Please help! You can try managing your time by prioritizing by importance and / or deadline. You can start attending to requirements that are due at an earlier date. It also helps if you regularly \emp{study} as compared to studying (or craming) for a test the night before, because you are always prepared. It will also be a good idea to start right away when projects are given, as also not to cram. You'll feel less stressed. There is no best time \emp{management} for everyone. Each one has his own time \emp{management} technique, so try and find your own style that works best for you. I hope this \emp{helps}! Good luck!} \\
\midrule
Adversarial & \tabincell{c}{Business \& Finance \\ Confidence: 50.0\%} & \multicolumn{1}{m{13cm}}{I need more time management \emp{competency}, can anyone help? I'm a high school student who is also \emp{fact} another course but I can't find enough time to get \emp{everything} done, I'm a week or two behind already. Please help! You can try managing your time by prioritizing by importance and / or deadline. You can start attending to requirements that are due at an earlier date. It also helps if you regularly \emp{examining} as compared to studying (or craming) for a test the night before, because you are always prepared. It will also be a good idea to start right away when projects are given, as also not to cram. You'll feel less stressed. There is no best time \emp{admin} for everyone . each one has his own time \emp{admin} technique, so try and find your own style that works best for you. I hope this \emp{helping}! Good luck!} \\
\bottomrule
\end{tabular}
}
\vspace{-0.3em}
\caption{The adversarial examples generated by FGPM on \textit{Yahoo! Answers} using Bi-LSTM model.}
\vspace{-1.8em}
\label{tab:adversarial_examples_on_bilstm}
\end{table*}

\end{document}